# Deep learning for fast segmentation and critical dimension metrology & characterization enabling AR/VR design and fabrication

Kundan Chaudhary[1], Subhei Shaar[1], Raja Muthinti[1]
[1]Meta Reality Labs

*Correspondence to: gmuthinti@meta.com

## Abstract

Quantitative analysis of microscopy images is essential in the design and fabrication of components used in augmented reality/virtual reality (AR/VR) modules. However, segmenting regions of interest (ROIs) from these complex images and extracting critical dimensions (CDs) requires novel techniques, such as deep learning models which are key for actionable decisions on process, material and device optimization. In this study, we report on the fine-tuning of a pre-trained 'Segment Anything Model' (SAM) using a diverse dataset of electron microscopy images. We employed methods such as low-rank adaptation (LoRA) to reduce training time and enhance the accuracy of ROI extraction. The model's ability to generalize to unseen images facilitates zero-shot learning and supports a CD extraction model that precisely extracts CDs from the segmented ROIs. We demonstrate the accurate extraction of binary images from cross-sectional images of surface relief gratings (SRGs) and Fresnel lenses in both single and multiclass modes. Furthermore, these binary images are used to identify transition points, aiding in the extraction of relevant CDs. The combined use of the fine-tuned segmentation model and the CD extraction model offers substantial advantages to various industrial applications by enhancing analytical capabilities, time to data and insights, and optimizing manufacturing processes.

## Introduction

The exponential growth in the volume of image data within device design and fabrication settings generated by electron microscopy, along with the increasing complexity of datasets, presents challenges, particularly in the manual analysis of images[1-3]. Extracting CDs such as etch depth, periodicity, slant angle, and thickness from ROIs using traditional methods has shown limitations[4], prompting the exploration of image segmentation as a promising solution[5-7]. Segmentation of ROIs is the initial step, particularly when assessing SRG[8] CDs and contours. This process is pivotal in improving the cycle of learning (CoL) in closed loop analysis (CLA) studies, which are essential for refining optical design and fabrication iterations[9]. The segmentation step extracts ROIs as binary images, which then allows for further quantitative analysis of the ROIs. However, manual delineation of ROIs is inherently subjective, requires a considerable amount of time, and can lead to inconsistent results. Automated segmentation offers a quicker, more accurate, and consistent solution, though it faces challenges due to the diversity in shape, size, and material composition of the ROIs, as well as the presence of ambiguous boundaries[10]. The variation is so substantial that it necessitates a segmentation model capable of generalizing across these differences.

Image segmentation presents inherent challenges as it requires the identification of features that vary spatially while preserving the contextual integrity of each pixel to ensure accurate semantic labeling[11]. Various methods have been in use to tackle image segmentation challenges before the advent of deep learning. These methods include thresholding[12], edge detection[13], histogram-based methods[14], clustering[15], region-growing methods[16], graph-based approaches[17], and watershed transformations[18]. The impact of deep learning on image segmentation has been notable[19]. Unlike traditional machine learning methods that depend on manually crafted features, demanding extensive domain expertise, convolutional neural networks (CNNs) autonomously learn relevant feature representations through comprehensive end-to-end training[20-22]. This capability not only eliminates the need for manual feature engineering but also enables the model to adjust and improve based on its predictions, offering a flexibility not afforded by traditional algorithms[23]. The use of transfer learning for electron microscopy datasets holds considerable promise, especially considering the advanced deep learning techniques necessary for successful image segmentation[24,25]. Furthermore, use of techniques such as LoRA allows for effective fine-tuning of downstream tasks without the need to retrain the entire model[26]. This step aids in developing effective segmentation models, which support precise contour detection and accurate CD extraction.

CNN-based techniques like Mask-RCNN are capable but come with their own set of challenges[27]. Specifically, they are computationally intensive and require a substantial amount of labeled data, which is particularly demanding when analyzing electron microscopy images. These images are particularly challenging to handle due to their high variability and the presence of imaging artifacts[10,28]. The need for large amounts of data for fine-tuning is circumvented by transformer-based architectures such as SAM, which is trained on a large and diverse set of one billion masks and displays superior performance across various metrics[29]. Besides, a qualitatively better mask prediction with crisper boundaries addresses the bottleneck in CD extraction accuracy.

In this work, we leverage transfer learning to fine-tune SAM[30-32], specifically targeting the extraction of ROIs from cross-sectional images of SRGs and Fresnel lenses. This approach results in more robust predictions. Here, the image dataset comprises scanning electron microscopy (SEM), transmission electron microscopy (TEM), and scanning transmission electron microscopy (STEM)-generated cross-sectional images of AR/VR devices at different processing steps of device fabrication. These images play a fundamental role in understanding the current stage of device fabrication and in improving the original design and processing steps. We emphasize SEM images in this work due to its widespread application. The number of images required to train the model is minimal compared to the image dataset needed for certain architectures such as Mask R-CNN. Our rigorous evaluation yielded a high intersection over union (IoU) scores for various layers, such as SRGs, substrate, atomic layer deposition (ALD), anti-reflective coating (ARC), and overcoat (OC). This underscores the importance of fine-tuned models for ROI segmentation. Additionally, we validate that upon accurate extraction of segmented ROIs, we can calculate relevant CDs that closely match the manual measurements.

**Results**

Figure 1a introduces an approach using a transformer-based architecture[33], inherent in SAM, that has been fine-tuned on AR/VR device data. This architecture accelerates the fine-tuning process through a LoRA-based technique, showing an improved performance compared to CNN-based architectures. The fine-tuned model processes a cross-sectional SRG image and generates a binary image, representing the ROI in white and the background in black. The contour of the ROI is overlaid on the original image to visually assess the quality of the ROI extraction, in addition to evaluating it with the IoU score of the model. The binary image facilitates the determination of transition points, such as the top and bottom regions of images, which are indispensable for effective CD extraction. Figure 1b illustrates the method for finding transition points, where a transition point is defined as a point with a pixel value of 255 (in an 8-bit image) and its neighboring pixels (left, right, top, or bottom) with a value of 0. Here, the transition points are determined based on the type of CD to be extracted. For example, when calculating the etch depth of the SRG, which represents the distance between the top (point 1 in Figure 1b) and the bottom (point 6 in Figure 1b) of the SRG, we obtain the absolute value of the difference between the y-coordinates of points 1 and 6. Similarly, to calculate the mid thickness of the SRG, we identify points 2 and 3, and calculate the absolute value of the difference between their x-coordinates. Additionally, periodicity can be calculated by identifying points 2 and 4, and calculating the absolute value of the difference between their x-coordinates. In this way, we can extract a set of CDs of interest on demand.

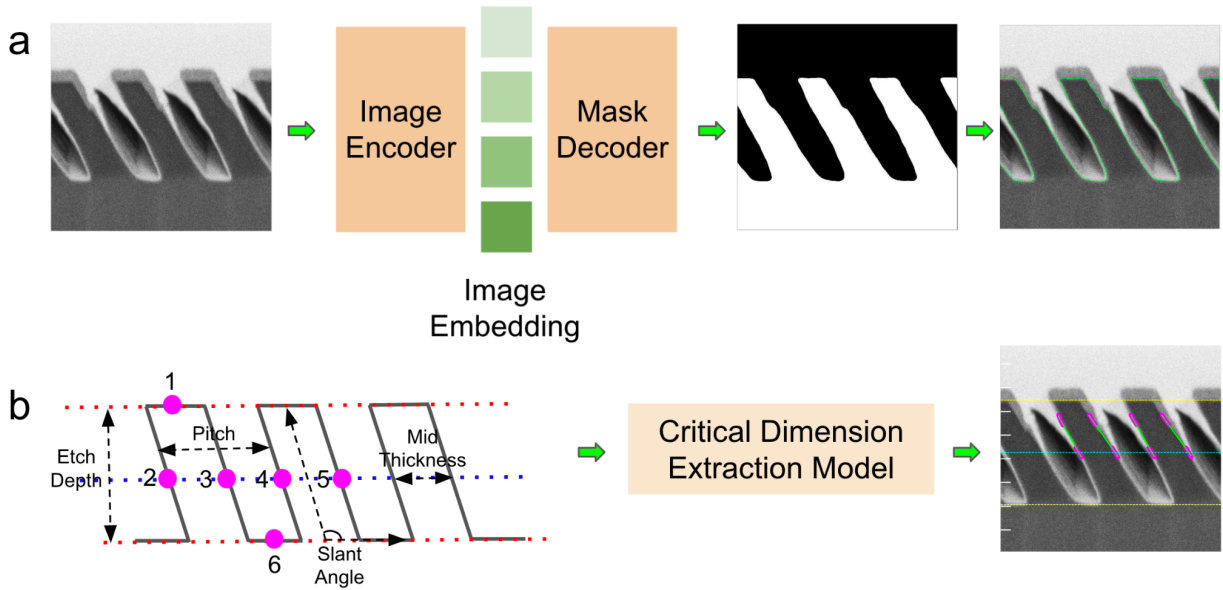

**Figure 1:** Segmentation and CD extraction from cross-sectional electron microscopy images. **(a)** Segmentation model training process. A foundation vision model such as SAM is used for fine-tuning on a range of cross-sectional electron microscopy images. The segmentation process generates a binary image where the ROI is white, and the background is black. The contour of the binary image is overlaid on the original image to visually assess the model's performance. **(b)** CD extraction process. After the segmentation step, a set of transition points are extracted from the binary image. Depending on the attribute of the CD, these transition points are grouped to define the CDs for a given repeat structure of the SRG. For calculating the horizontal thickness, x-coordinates of the points are relevant, while y-coordinates are used for the vertical thickness.

The effectiveness of the segmentation model is quantified using the IoU score, Figure 2a, a metric that measures the similarity between the binary masks produced by our fine-tuned model and the ground truth. This high level of similarity confirms the quality of the binary masks, which is essential for accurate CD extraction. As semSAM (fine-tuned SAM on electron microscopy images) outperforms all other fine-tuned state-of-the-art models, we selected it for extracting ROIs and analyzing their CDs. The interquartile range in the box plot represents the performance of the segmentation model in extracting a range of ROIs, including SRGs, ARC, OC, substrate, among others. Figure 2b-g showcases the model's ability to handle a diverse range of complexities, such as variations in component structures and image artifacts in SEM images of AR/VR devices. These complexities include presence of additional surfaces on top (Figure 2c), thin geometry with a high aspect ratio (Figure 2d), the presence of imaging artifacts (Figure 2e), imaging with a low electron dose (Figure 2f), and a fractured cross-section with low contrast between the ROI and background (Figure 2g). The ROI's contour, overlaid in red, illustrates how our model adeptly segments ROIs from backgrounds that vary in brightness, contrast, and noise—challenges commonly encountered with different electron microscopy systems. Notably, the model also applies to cross-sectional images generated by STEM (Figure 2h) and TEM (Figure 2i), beyond those from SEM. This technique-agnostic segmentation capability boosts research and production opportunities. It allows for the use of a range of imaging techniques to acquire images from the imaging tool that best serves the device processing step, providing context-specific solutions. Moreover, the generalized segmentation capability enables the segmentation of arbitrary geometries beyond SRG architectures, such as a Fresnel lens (Figure 2j) fabricated using the focused ion beam (FIB) technique. A Fresnel lens helps in confining light within a specific radiation cone of interest. Binary images, obtained from segmentation, are valuable for assessing the capabilities of the lens manufacturing process and guiding improvements in the design of these lenses. Here, each 2D cross-sectional SEM image of Fresnel lens undergoes segmentation, and the extracted contour points are then utilized to generate connected meshes, facilitating layer-to-layer connectivity. This scheme is very useful in extracting a 3D view of the Fresnel lens, which would otherwise be challenging due to the complexity of the images.

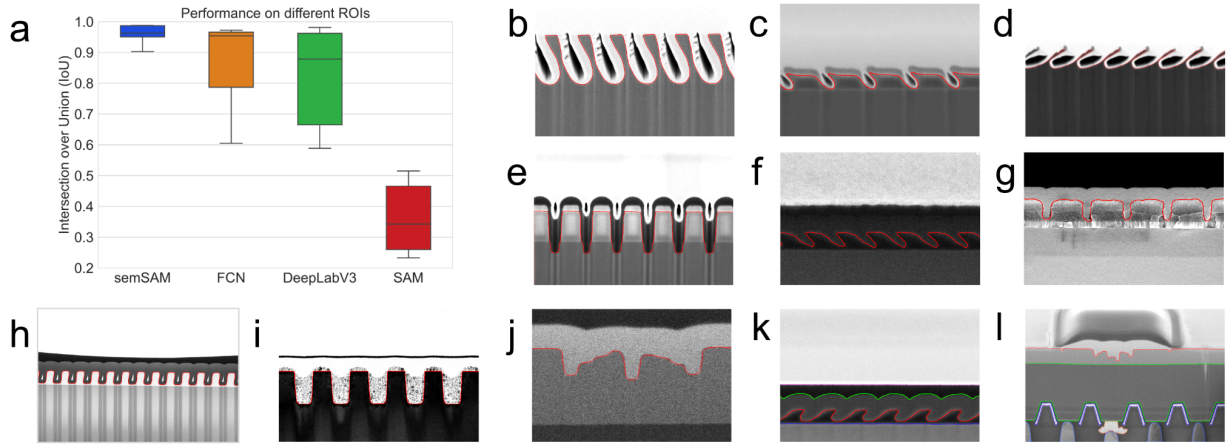

**Figure 2:** Segmenting cross-sectional electron microscopy images with a fine-tuned segmentation model. **(a)** Performance of different models fine-tuned on AR/VR data and across different ROIs, compared to vanilla SAM. The IoU score, which measures the accuracy of the predicted ROIs against the actual ones, indicates that a higher IoU score corresponds to better model performance in segmentation tasks. **(b)** Segmentation of a simple geometry, showing good contrast between it and the background. The segmented ROI is highlighted with a red contour for visual quality assessment. **(c)** SRG with some material deposited on top. **(d)** High aspect ratio SRG. **(e)** SRG with both processing and imaging artifacts. **(f)** SRG imaged with a low electron dose, resulting in a noisier image and reduced contrast with the background. **(g)** SRG with low contrast against the background. Segmentation performance on SRG cross-sections imaged with **(h)** STEM and **(i)** TEM. **(j)** Segmentation of a cross-sectional SEM image of a Fresnel lens. Multiclass segmentation reveals the model's capability to **(k)** segment different geometries within an SRG architecture and **(l)** μLED.

Figure 2k-l emphasizes the model's versatility in segmenting multiple classes within a single SEM image, which is particularly beneficial for analyzing complex optical elements like waveguides[34]. Figure 2k highlights how different structures within a waveguide, each contributing to optical performance, are segmented with high degrees of success. The variation in segmentation is quantified by the interquartile range of IoU scores in Figure 2a, which reflects their geometric complexity. Likewise, the model is successful at extracting various ROIs of interest from a μLED stack (Figure 2l). This capability allows for a quantitative understanding of the fabricated sample and identifies potential steps that can be taken to achieve a desirable device.

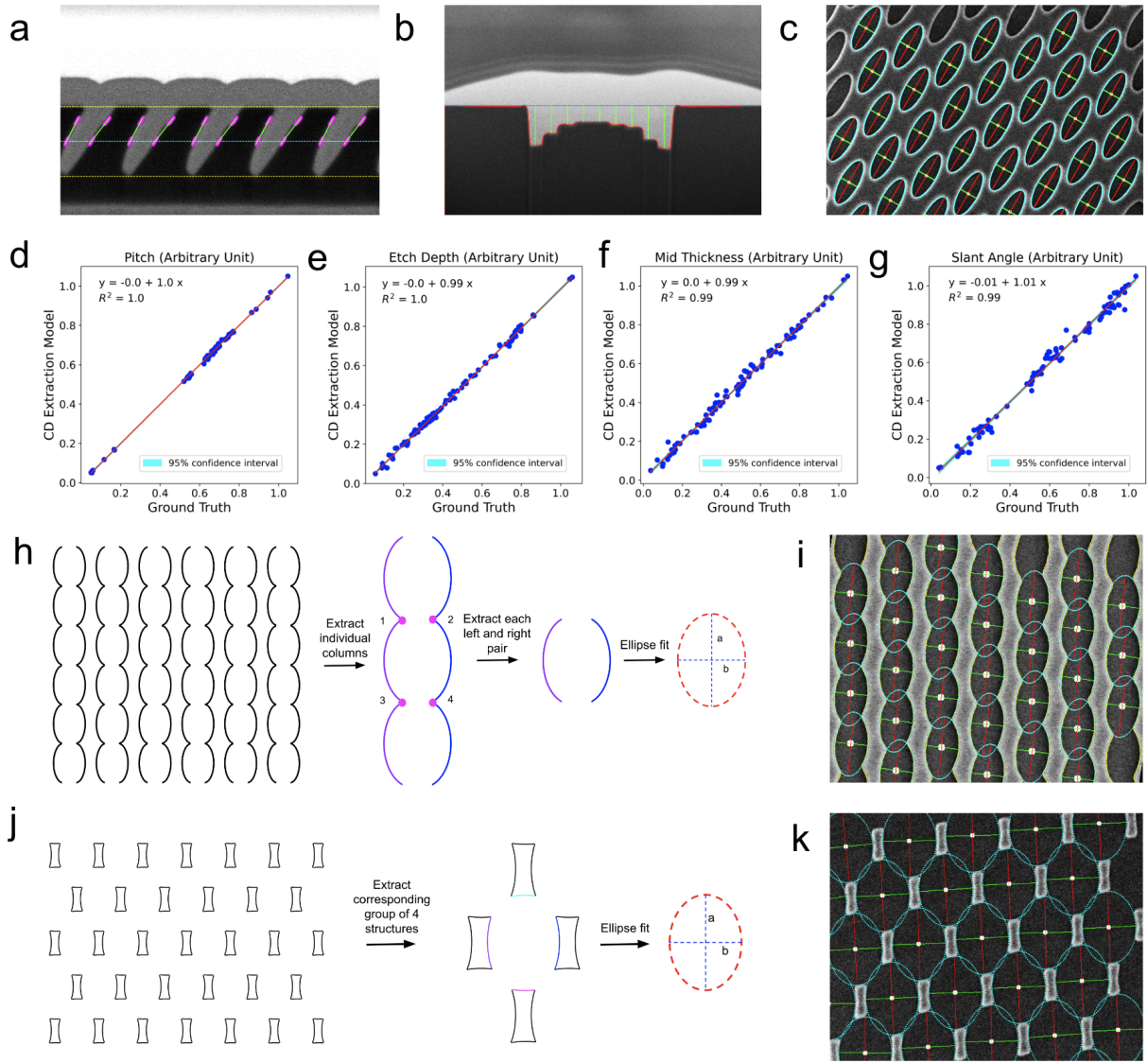

**Figure 3:** CD extraction from segmented ROIs. **(a)** CD extraction of an SRG involves determining the top and bottom edges to calculate the etch depth. Transition points along the middle of the top and bottom are used to calculate the mid thickness of each SRG repeat unit. A similar strategy helps calculate the periodicity of the SRGs. Transition points near the top-left and mid-left are used to calculate the angle along the left side-wall, while points near the top-right and mid-right are used to calculate the angle along the right side-wall. **(b)** For the Fresnel lens, the depth of individual 9 bins is calculated by the difference between the top region, shown by the blue horizontal line, and the bottom region of each bin. **(c)** CD extraction of a 2D grating consisting of elliptical units. After segmenting the inner ellipse, parameters such as major/minor axes and angle of rotation are extracted by fitting the contour with an ellipse. Comparison of CDs of interest, such as **(d)** pitch, **(e)** etch depth, **(f)** mid thickness, and **(g)** slant angle, manually measured versus those extracted by the CD extraction model. Correlation plots show a strong agreement between manually measured ground truth and model extracted CDs. **(h)** Schematic representation of merged ellipses forming a columnar architecture. A pair consisting of the left and right side of the potential ellipse is extracted and fitted with an ellipse to obtain ellipse parameters. **(i)** Extracting ellipse parameters of merged ellipses. **(j)** Schematic representation of islands. Four neighboring structures are used to extract relevant points of an expected ellipse and thereafter fitted with an ellipse to obtain ellipse parameters. **(k)** Extracting ellipse parameters of heavily merged ellipses forming islands.

Figure 3 depicts the capabilities of our model in extracting CDs from cross-sectional SEM images. By accurately identifying transition points within binary segments, the CD extraction model can compute various CDs, including thickness, width, and periodicity of an SRG. This technique is applicable not only to SRG structures (Figure 3a) but also to more complex forms like Fresnel lens, where it can extract depths of various bins across a Fresnel lens (Figure 3b), and 2D gratings containing elliptical units (Figure 3c). The extracted CDs of SRG structures, such as pitch (Figure 3d), etch depth (Figure 3e), mid thickness (Figure 3f), and slant angle (Figure 3g), closely match the ground truth (i.e., manual measurements). This supports confidence in using such CD extraction models, paving the way for a method that is more accurate, consistent, and quick for extracting CDs compared to the tedious, time consuming, and inconsistent manual methods. Figure 3h addresses the challenges associated with extracting CDs from complex 2D grating structures, which consist of a repeating ellipse as shown in Figure 3c, but in a merged condition. Here, during the processing step, the individual ellipses merge with each other to form a columnar structure. The fine-tuned segmentation model is used to extract these columns, and the left and right contours of each column are fitted separately with a cosine function in order to obtain the crest and troughs on each side. The points from one crest to another on the left side are paired with the points from one trough to another on the right side, such that they form the left and right sides of a potential ellipse. These points are further fitted to an ellipse to obtain parameters such as the major and minor axes (Figure 3i). The case of ellipses merging heavily, starting from isolated cases as in Figure 3c, is illustrated in Figure 3j, where a periodic dog-bone-like structure is formed. To extract the ellipse parameters, the fine-tuned segmentation model is used to obtain the binary image of dog-bone-like structures, which are then grouped into four-member groups. Here, the left structure's right side (purple arc in Figure 3j) contributes to the left side of the potential ellipse, while the top structure's bottom side (cyan arc) contributes to the top side, the right structure's left side (blue arc) contributes to the right side, and the bottom structure's top side (magenta arc) contributes to the bottom side. Subsequently, these points from the left, top, right, and bottom sides are fitted to an ellipse to extract its parameters, as shown in Figure 3k. As a whole, Figure 3 showcases the model's ability to handle challenging segmentation tasks, providing detailed insights that are imperative for advancing the design and fabrication of optical elements.

| Attributes | Classical Methods (Edge Detection, Thresholding) | Deep Learning (Fine-Tuned Foundation Vision Models) |
|---|---|---|
| User Expertise Required | High; requires knowledge in image processing techniques | Moderate; relies more on data and less on domain-specific expertise |
| Scalability | Limited scalability due to manual tuning and adjustments | Highly scalable with automated processes and model reusability |
| Flexibility in Handling Variability | Low flexibility; struggles with variations in image quality and content | High flexibility; robust to changes in image quality and content |
| New Application Development Time | Can take more than a week | Less than a day |
| Success Rate | >50% and varies depending upon images and ROIs | >95% |

**Table 1:** Comparison of classical image processing and deep learning based methods for CD extraction.

**Discussion**

This study highlights the efficacy of deep learning techniques, particularly the fine-tuned SAM model, in segmenting complex image datasets derived from AR/VR device manufacturing. The proposed method addresses the challenge of accurately extracting CDs from ROIs in images produced by electron microscopy. This approach augments the accuracy and efficiency of ROI analysis. Additionally, it improves the automation and precision of CD extraction in SEM images. The model's ability to autonomously adapt to new segments enhances its utility, making it a useful tool in industries where precise extraction of CDs is essential.

We focused on SRGs and Fresnel lenses because of their intricate structures, which pose ample segmentation challenges. These components are key in developing AR/VR devices. The model can be swiftly updated with new data, which allows for faster annotation and retraining. This enriches its performance on unseen images. Notably, the CD extraction algorithm remains consistent, yet it can be adapted to extract additional measurements as needed. The model's performance is independent of the electron microscope type used for image capture.

The integral role of supervised methods in precise image segmentation, particularly in fields like semiconductor processing and AR/VR devices fabrication, cannot be overstated. These methods pave the way for scalable applications where the model, once fine-tuned, operates independently of image processing experts (Table 1). By simply integrating a few representative data samples from the device, which capture the expected variations in imaging quality and content, the model can be updated to maintain high accuracy across different manufacturing scenarios. This generalization capabilities of SAM-like models and their application in core manufacturing processes underscore the vital role of advanced image segmentation in modern industrial applications. This drives innovation and improvements in product design and quality.

Looking forward, the potential of this method to impact fields such as semiconductor manufacturing, materials science, and biomedical imaging is considerable. Future work will aim to further refine the model's generalization abilities and explore its application in complex imaging scenarios beyond AR/VR devices. This study establishes a new benchmark for image segmentation and CD extraction in industrial applications, paving the way for more sophisticated, automated, and precise manufacturing processes.

**Methods**

Image Dataset Generation
SEM/TEM/STEM tools were utilized to obtain cross-sectional images of various device designs across a range of processing steps.

Image Annotation
Besides the use of custom scripts involving image thresholding and object detection methods, open source Labelme (https://github.com/labelmeai/labelme) software was used for annotating images.

## Author Information

Authors and Affiliations

Meta Reality Labs, Redmond, WA, 98052
Kundan Chaudhary, Subhei Shaar, and Raja Muthinti

## Contributions

K.C. conceived the project. S.S. generated image datasets. K.C. performed calculations, developed the critical dimension extraction model, and analyzed data. R.M. led the project. All authors contributed to discussions and manuscript preparation.

## Corresponding Authors

Correspondence to [Raja Muthinti](#).